\documentclass[11pt,letterpaper]{article}
\usepackage[T1]{fontenc}
\usepackage{lmodern}
\usepackage[margin=0.9in]{geometry}
\usepackage{microtype}
\usepackage{amsmath,amssymb}
\usepackage{graphicx,placeins}
\usepackage{array,booktabs,calc,multirow}
\usepackage[font=small,labelfont=bf]{caption}
\usepackage{textcomp}
\usepackage{xurl}
\usepackage[hidelinks]{hyperref}

\newcommand{\ul}[1]{\underline{#1}}
\providecommand{\doi}[1]{\url{https://doi.org/#1}}

\begin{document}
\title{COMPASS: Cross-modal Observation with Multi-Path Adaptive Stopping and Selection for Early Time Series Classification}
\author{%
  Cheng-En Chung\textsuperscript{\(1\)}\quad
  Yujia Wu\textsuperscript{\(1\)}\quad
  Oscar Tai-Yuan Chen\textsuperscript{\(1\)}\quad
  Vincent S. Tseng\textsuperscript{\(1\,\dagger\)}\\[0.5em]
  \small \textsuperscript{\(1\)}Department of Computer Science, National Yang Ming Chiao Tung University, Taiwan\\[0.25em]
  \small \textsuperscript{\(\dagger\)}Corresponding author:
  \href{mailto:vtseng@cs.nycu.edu.tw}{\nolinkurl{vtseng@cs.nycu.edu.tw}}%
}
\date{}
\hypersetup{
  pdftitle={COMPASS: Cross-modal Observation with Multi-Path Adaptive Stopping and Selection for Early Time Series Classification},
  pdfauthor={Anonymous Authors}
}
\maketitle

\begin{abstract}
Early time-series classification (ETSC) balances predictive performance against the amount of observed data. In multimodal settings, modality-associated and fused branches may differ in predictive ability at the same observed prefix, motivating joint decisions about when to stop and which branch should provide the final prediction. We propose COMPASS, a two-stage framework that preserves two modality-associated temporal branches and one fusion branch. The first stage combines cross-modal feature exchange and gated fusion with branch-wise supervision at every prefix. The second stage optimizes a shared four-action controller to continue observation or terminate through one of the frozen branches. Experiments on six physiological and wearable datasets compare COMPASS with multimodal adaptations of five ETSC baselines. COMPASS achieves the best average ranks in accuracy-earliness harmonic mean, AUROC, and earliness within this comparison set. At identical learned stopping times, branch selection improves mean accuracy over fusion readout on five of the six datasets, with gains of 2.35 and 2.16 percentage points on ISRUC-Sleep and UCDDB, respectively. To the best of our knowledge, this is the first work on joint stopping and branch selection for multimodal early time series classification.

\end{abstract}

\section{Introduction}\label{introduction}

Early time series classification (ETSC) aims to predict a time
series label from partial observations while balancing predictive
performance against the cost of waiting \cite{renault2025survey}. This trade-off is
critical in physiological monitoring, where earlier decisions enable
timely intervention. Such monitoring often involves multiple modalities,
such as electrocardiography (ECG) paired with respiration or
photoplethysmography (PPG), which provide complementary and partly
redundant information about the same physiological state \cite{tejedor2019heartbeat,john2026wearable}. These modalities may also become informative at different
times, so the evidence available for a decision accumulates
asynchronously. Exploiting this evidence requires both effective
multimodal representations and a decision mechanism that determines when
the available evidence is sufficient for classification.

Multimodal early classification has been studied in several
settings. In remote sensing, optical and radar time series have been
fused for in-season crop classification \cite{weilandt2023crop}, and recent work
further incorporates calibrated acceptance and deferral decisions
\cite{han2026sar}. Cao et al. combine heterogeneous sequence representations
with a stopping policy supervised by prefix-wise classification losses
and time costs \cite{cao2023early}. For multivariate time series, CHARLEE
adapts channel usage together with stopping to reduce input utilization
\cite{pantiskas2023charlee}. Among the methods that make stopping or deferral
decisions \cite{han2026sar,cao2023early,pantiskas2023charlee}, the label is produced by a
predetermined prediction head. These methods determine when to stop or
which input to use, but not which branch produces the final label.

This distinction matters when evidence accumulates asynchronously. At a given prefix, one modality may already be discriminative while the other remains uninformative. A fused branch can exploit cross-modal interactions, but its prediction need not be more accurate than that of a modality-associated branch at every prefix. More generally, jointly trained multimodal models can
underperform their best unimodal counterparts \cite{wang2020multimodal}. A prediction
branch anchored to a single modality provides an alternative
readout that depends less on the other modality. Their relative
predictive utility may therefore differ across samples and observation
stages, so acquiring both modalities up to the same stopping time does
not guarantee that the fused branch yields the preferable prediction for
every sample. The question differs from deciding which inputs to acquire
\cite{pantiskas2023charlee}: it concerns which branch should provide the label from the information already observed. We address joint stopping and branch selection in multimodal ETSC, where an early decision specifies both whether to continue observing and, upon termination, which available branch provides the prediction.

We propose COMPASS, a two-stage framework that incorporates branch selection into the early decision itself. In the first stage, two modality-associated temporal branches and one fusion branch are trained with separate classification supervision at every prefix. The modality-associated branches retrain features prior to cross-modal exchange, while the fusion pathway combines the exchanged features via gated attention. This design preserves alternative prediction pathways alongside cross-modal integration. In the second stage, a single controller receives the hidden states of all three frozen branches and selects among four actions: continue observing, stop with the fusion branch, or stop with either of the modality-associated branches. Each terminal action specifies both the stopping time and the branch that produces the final label. Whereas gated fusion softly reweights modalities within a fused representation, the controller makes a discrete choice among complete prediction branches.

We evaluate COMPASS on six physiological and wearable datasets
against multimodal adaptations of five ETSC baseline families. Within
this comparison set, COMPASS achieves the best mean ranks in
Harmonic Mean Accuracy (HM-ACC), AUROC, and earliness. We further
distinguish overall performance from the contribution of terminal branch
choice by comparing routed and fixed-fusion predictions at identical
learned stopping points. Matched-stop analysis (Section~\ref{matched-stop-analysis-of-branch-selection}) shows positive mean accuracy differences relative to
fixed-fusion prediction across five of the six datasets, with the
magnitude and direction of routing effects varying across datasets. The
results support examining branch selection explicitly rather than
inferring its benefit from aggregate classification-earliness scores
alone.

The main contributions of this work are summarized as follows:

1. \textbf{Joint stopping and branch selection.} We formulate multimodal ETSC as a joint decision over observation time and terminal prediction source. A shared four-action policy chooses between continued observation and termination through a modality-associated or fusion branch, making branch selection part of the stopping action.

2. \textbf{Branch-preserving multimodal representation learning.} We
integrate cross-modal feature exchange and gated fusion with separately
supervised temporal branches, preserving modality-associated and
fused prediction pathways and enabling the controller to
assess the evidence at each observed prefix.

3. \textbf{An evaluation separating overall performance from terminal
branch-selection effects.} We evaluate COMPASS on six physiological
and wearable datasets through baseline comparisons, alternative stopping
configurations, and architectural ablations. Matched-stop rescue
and harm analysis compares routed and fusion predictions at identical
learned stopping times, quantifying the benefit and cost of branch
selection while holding the observed prefixes fixed.

To the best of our knowledge, this is the first work on joint stopping and branch selection for multimodal early time series classification.

\section{Related Work}\label{related-work}

\textbf{Early time-series classification (ETSC).} ETSC balances prediction quality against the cost of continued
observation \cite{renault2025survey}. Existing approaches differ primarily in how
they determine when to stop. TEASER evaluates the reliability of prefix
predictions \cite{schafer2020teaser}, whereas EARLIEST learns an adaptive halting
policy \cite{hartvigsen2019earliest}. Snippet-based methods (SPN) make sequential
decisions based on local observations; SPN-V2 further employs
knee-guided neuroevolution to optimize the accuracy-earliness trade-off
for multi-lead ECG classification \cite{huang2024spnv2,huang2023spn}. ELECTS learns
classification and stopping behavior within an end-to-end framework
\cite{russwurm2023elects}, while CALIMERA and UTSC investigate cost-based and
uncertainty-aware decisions, respectively \cite{bilski2023calimera,yan2025utsc}. These studies
provide the foundations for learning when sufficient evidence has
accumulated. CHARLEE \cite{pantiskas2023charlee} extends early classification of
multivariate time series through adaptive channel filtering and
stopping, balancing classification accuracy against input savings across
time and channels. COMPASS instead maintains a fixed set of
observed modalities and jointly selects the stopping time and the
terminal prediction
branch.

\textbf{Multimodal representation learning.} Multimodal learning combines complementary information through
interactions at the feature, representation, or prediction level
\cite{baltrusaitis2019survey}. MMTM enables cross-modal feature recalibration through
channel-wise modulation \cite{joze2020mmtm}. For physiological signals,
CoRe-Sleep coordinates unimodal and multimodal representation learning
to improve robustness to imperfect modalities \cite{kontras2024coresleep}. Such work
shows that modality-specific representations and cross-modal
interaction can coexist within a prediction framework. COMPASS builds on
these principles through cross-modal feature exchange, gated
fusion, and separately supervised temporal branches. Its contribution
therefore lies not in preserving multiple branches per se, but in using
them as candidate readouts for a terminal decision at a partially
observed
prefix.

\textbf{Multimodal early decisions.} Early
decision-making with multiple information sources has also been
studied directly. Cao et al. \cite{cao2023early} process sequentially
arriving multimodal elements through modality-specific encoders and a
shared fusion architecture, with a stopping policy supervised by
prefix-wise classification losses and time costs. In remote sensing,
Weilandt et al. \cite{weilandt2023crop} fuse radar and optical time series using
temporal attention for in-season crop classification, and Han et
al. \cite{han2026sar} address asynchronous radar-optimal fusion by
calibrating and thresholding predictions for acceptance or deferral.
In both \cite{cao2023early} and \cite{han2026sar}, the label is produced by a
predetermined prediction head: the stopping or deferral decision
determines when a prediction is released, but not which branch
produces
it.

COMPASS addresses this choice by selecting the prediction source
among modality-associated and fused branches at a common observed
prefix. Its single controller couples this choice with stopping, and the
matched-stop fixed-fusion readout control isolates the accuracy
changes attributable to terminal branch selection at the learned
stopping times. Our numerical benchmark adapts five ETSC baseline
families, namely TEASER, ELECTS, SPN\_V2, CALIMERA, and UTSC, to the
evaluated physiological and wearable datasets under a common
temporal stopping protocol (Section~\ref{experimental-setup}).

\section{Problem Formulation}\label{problem-formulation}

We consider multimodal ETSC with two modalities recorded on a common
time axis. Let
\(\mathcal{D} = \{\left( \mathbf{X}_{i},y_{i} \right)\}_{i = 1}^{N}\)
denote a binary-labeled dataset, where
\(\mathbf{X}_{i} = \left( X_{i}^{(1)},X_{i}^{(2)} \right)\) contains two
synchronized signals observed over a window of duration \(T_{i}\), and
\(y_{i} \in \{ 0,1\}\), with 1 denoting the positive class. Sample
\(i\) has \(K_{i}\) decision points. At decision point
\(k \in \{ 1,\ldots,K_{i}\}\), the model can access only the prefixes
ending at \(t_{i,k} \leq T_{i}\):

\begin{equation}
\mathcal{P}_{i,k} = \left( X_{i}^{(1)}\left\lbrack 0:t_{i,k} \right\rbrack,\mspace{6mu} X_{i}^{(2)}\left\lbrack 0:t_{i,k} \right\rbrack \right).
\end{equation}

Both modalities are observed up to the same time, but their predictive
utility may differ at a given prefix.

The model maintains three prediction branches, indexed by
\(\mathcal{B} = \left\{ F,1,2 \right\}\), corresponding to Fusion, MD1,
and MD2. Each branch \(b\) outputs a class distribution
\(\mathbf{p}_{i,k}^{(b)}\), whose positive class entry
\(p_{i,k}^{(b)}(y = 1) \in \lbrack 0,1\rbrack\) serve as its prediction
score. MD1 and MD2 retain modality-associated pathways using
features before cross-modal exchange, while Fusion combines information
from both modalities.

At each decision point, a single controller either requests
the next synchronized snippets or terminates through a selected branch.
Its actions are Continue (\(0\)), Stop-Fusion (\(1\)), Stop-MD1 (\(2\)),
and Stop-MD2 (\(3\)):

\begin{equation}
\mathcal{A} = \{ 0,1,2,3\}.
\end{equation}

Each action depends only on the observed prefix. Let \(k_{i}^{*}\) be
the first decision point at which a stopping action is selected, and
\({\widehat{b}}_{i}\) its associated branch. If the controller continues
through the final decision point, it terminates at \(K_{i}\) and uses
Fusion. The final prediction is
\({\widehat{y}}_{i}= \mathbb{1}\lbrack p_{i,k_{i}^{*}}^{\left( {\widehat{b}}_{i} \right)}(y = 1) \geq \gamma^{*}\rbrack\),
where \(\gamma^{*}\) is determined from training data and retained for
testing. The stopping time is \(\tau_{i} = t_{i,k_{i}^{*}}\), and the
observed fraction is \(e_{i} = \tau_{i}/T_{i}\).

The objective is to balance classification performance, measured by
accuracy and AUROC, against earliness
\(EAR = \frac{1}{N}\sum_{i = 1}^{N}e_{i}\), by learning both when to
stop and which branch supplies the final prediction. Both modalities
remain available until the stopping time; branch selection determines
the terminal prediction source rather than the sensor acquisition
schedule.

\section{Method}\label{method}

\begin{figure}[!htbp]
\centering
\includegraphics[width=\linewidth]{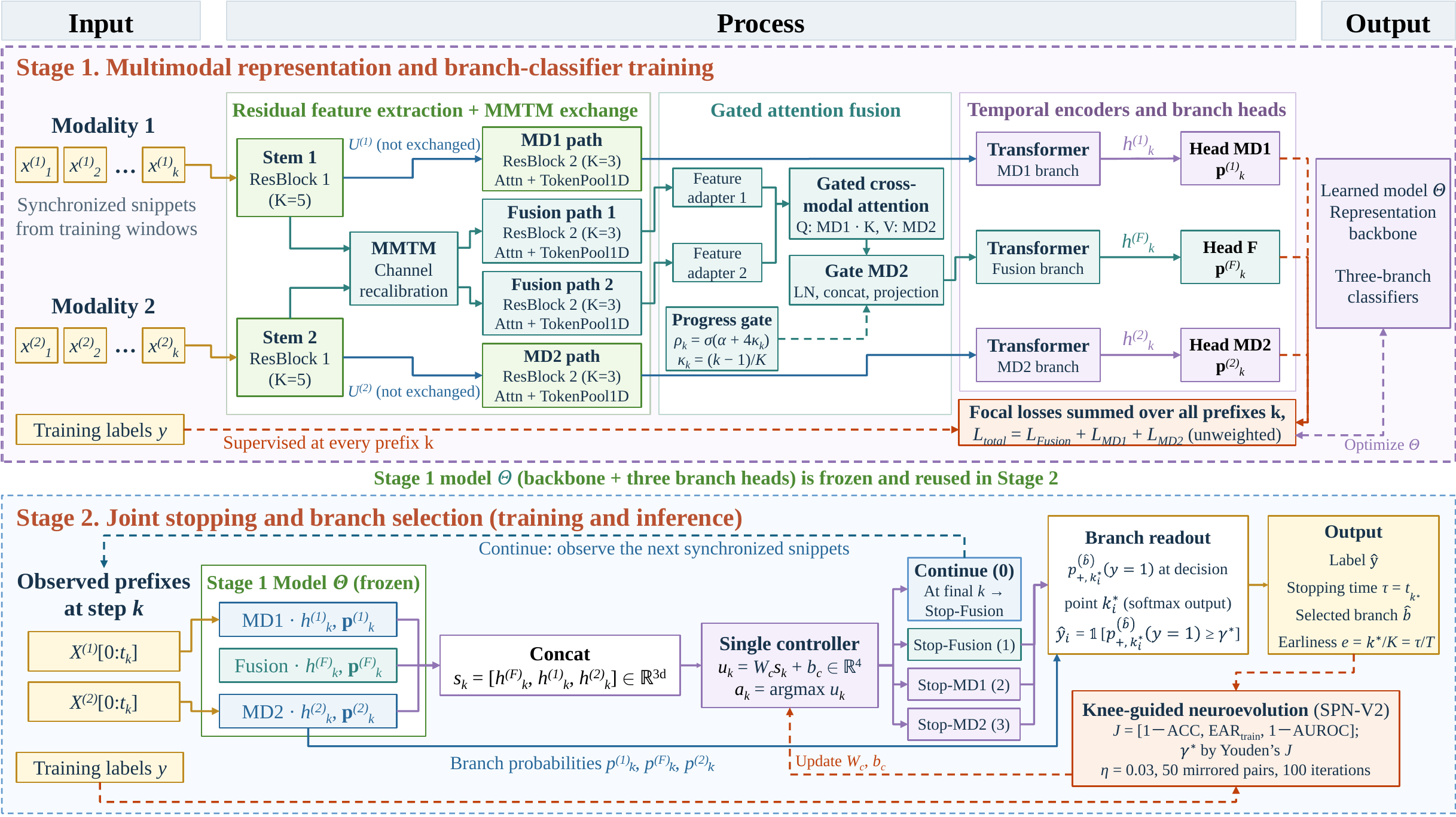}
\caption{Overall architecture of COMPASS.}\label{fig:architecture}
\end{figure}

COMPASS makes the prediction source part of the early-stopping
decision. As illustrated in Figure~\ref{fig:architecture}, Stage 1 learns two
modality-associated prediction branches and a fusion branch, with
supervision at every observed prefix. Stage 2 freezes these
representations and classifiers and learns a single controller that
chooses between continued observation and termination through one of the
three branches. This separation exposes alternative prediction pathways
while allowing the terminal decision to account for their joint state.

\subsection{Branch-Preserving Feature
Extraction}\label{branch-preserving-feature-extraction}

For sample \(i\), let \(x_{i,k}^{(m)}\) denote the \(k\)-th
synchronized snippet from modality \(m \in \{ 1,2\}\), where
\(k = 1,\ldots,K_{i}\). Modality-specific convolutional stem first
extracts local features:

\begin{equation}
\mathbf{U}_{i,k}^{(m)} = S_{m}\left( \mathbf{x}_{i,k}^{(m)} \right)
\end{equation}

The feature extractors use residual convolutional blocks \cite{he2016resnet}
with kernel sizes of 5 and 3 at the first and second residual
stages, respectively. An MMTM module \cite{joze2020mmtm} recalibrates channel
responses using information from both modalities:

\begin{equation}
\left( {\widetilde{\mathbf{U}}}_{i,k}^{(1)},{\widetilde{\mathbf{U}}}_{i,k}^{(2)} \right) = {MMTM}\left( \mathbf{U}_{i,k}^{(1)},\mathbf{U}_{i,k}^{(2)} \right)
\end{equation}

The backbone maintains separate pathways for the original and
recalibrated features. The MD pathways receive the features before
MMTM exchange, whereas the fusion paths receive the recalibrated
features:

\begin{align}
\mathbf{E}_{i,k}^{(m)} & = P_{m}^{MD}\left( \mathbf{U}_{i,k}^{(m)} \right) \\
\mathbf{V}_{i,k}^{(m)} & = P_{m}^{F}\left( {\widetilde{\mathbf{U}}}_{i,k}^{(m)} \right)
\end{align}

Each pathway includes the corresponding second residual stage and
attention-based token pooling. The resulting token representations
retain the local information needed for temporal encoding. Feature
adapters then map the fusion-path representations into a common feature
space:

\begin{equation}
\mathbf{A}_{i,k}^{(m)} = {Adapter}_{m}\left( \mathbf{V}_{i,k}^{(m)} \right)
\end{equation}

This organization allows cross-modal interaction to support fusion
while retaining unexchanged features for the two modality-associated
classifiers. The four paths thus feed three prediction branches:
the MD1 and MD2 paths from the two modality-associated
branches, and the two fusion paths are combined into the
fusion branch (Section~\ref{progress-aware-gated-attention-fusion}). Because each stem \(S_{m}\)
also feeds the exchange path and is trained jointly with the fusion
branch, we refer to the MD branches as modality-associated rather
than modality-specific. The three branches are supervised
separately and remain available to the stopping controller.

\subsection{Progress-Aware Gated Attention
Fusion}\label{progress-aware-gated-attention-fusion}

The fusion module asymmetrically combines the adapted token sequences:
modality 1 supplies the attention queries, while modality 2 supplies the
keys and values. For clarity, we omit the sample index \(i\) in
this subsection, so \(K\) denotes the number of decision points of
the current sample. Let
\(\mathbf{A}_{k}^{(m)} \in \mathbb{R}^{L \times d}\) denote the output
of feature adapter \(m\). The fusion module applies an additional
modality-specific projection:

\begin{align}
\mathbf{H}_{k}^{(m)} & = {GELU}\left( \mathbf{A}_{k}^{(m)}\mathbf{W}_{m} + \mathbf{b}_{m} \right) \\
\mathbf{C}_{k} & = {MHA}\left( \mathbf{H}_{k}^{(1)},\mathbf{H}_{k}^{(2)},\mathbf{H}_{k}^{(2)} \right)
\end{align}

Where \({MHA}(Q,K,V)\) denotes multi-head attention. The projected
modality-1 tokens and cross-modal context jointly determine a
feature-wise gate:

\begin{equation}
\mathbf{G}_{k} = \sigma\left( {LN}\left\lbrack \mathbf{H}_{k}^{(1)}\|\mathbf{C}_{k} \right\rbrack\mathbf{W}_{g} + \mathbf{b}_{g} \right)
\end{equation}

Here, \(\sigma\) is the sigmoid function, \(LN\) is layer normalization,
and \(\|\) denotes feature-wise concatenation. A scalar progress
gate additionally modulates the contribution of modality 2:

\begin{equation}
\kappa_{k} = \frac{k - 1}{K},\rho_{k} = \sigma\left( \alpha + 4\kappa_{k} \right)
\end{equation}

The scalar \(\alpha\) is learnable and initialized to \(- 4\); the
coefficient \(4\) is fixed. The progress gate starts from the first
observed snippet and increases with the decision index for a fixed
\(\alpha\). An initialization,
\(\rho_{1} = \sigma( - 4) \approx 0.018\), so modality 2 contributes
little to the earliest fused representations. The gate is broadcast
across tokens and feature channels.

The fused snippet representation is then constructed as:

\begin{align}
{\widehat{\mathbf{H}}}_{k}^{(2)} & = \rho_{k}\left( \mathbf{G}_{k} \odot \mathbf{H}_{k}^{(2)} \right) \\
\mathbf{E}_{k}^{(F)} & = {GELU}\left( {LN}\left\lbrack \mathbf{H}_{k}^{(1)}\|{\widehat{\mathbf{H}}}_{k}^{(2)} \right\rbrack\mathbf{W}_{f} + \mathbf{b}_{f} \right)
\end{align}

Where \(\odot\) denotes element-wise multiplication. The attention
context determines the feature-wise gate; the final fusion projection
combines modality-1 tokens with gated modality-2 tokens. Together, the
two gates make fusion depend on both the current cross-modal content and
the progress of observations.

\subsection{Temporal Branches and Prefix
Supervision}\label{temporal-branches-and-prefix-supervision}

Each branch \(b \in \mathcal{B} = \left\{ F,1,2 \right\}\) maintains its
own temporal encoder and classifier. At decision point \(k\), the
available snippet tokens are concatenated along the sequence dimension:

\begin{equation}
\mathbf{Z}_{i,k}^{(b)} = \mathcal{T}_{b}\left( {Concat}_{seq}\left\lbrack \mathbf{E}_{i,1}^{(b)},\ldots,\mathbf{E}_{i,k}^{(b)} \right\rbrack \right)
\end{equation}

The operator \(\mathcal{T}_{b}\) comprises a linear embedding,
positional encoding, a two-layer Transformer encoder with eight
attention heads \cite{vaswani2017attention}, and a linear output projection. Its input
contains only the token history available at the current boundary.
Attention can operate over that observed history without receiving
subsequent snippets.

The encoded sequence grows with the observed prefix. COMPASS
therefore uses a branch- and prefix-specific projection to obtain a
fixed-dimensional state:

\begin{equation}
\mathbf{h}_{i,k}^{(b)} = \mathbf{W}_{b,k}^{P}{vec}\left( \mathbf{Z}_{i,k}^{(b)} \right) + \mathbf{c}_{b,k}^{P} \in \mathbb{R}^{d}
\end{equation}

For \(L\) tokens per snippet and encoded token dimension \(d\), this
projection maps \(kLd\) inputs to \(d\) features. Separate projection
parameters are maintained for each branch and decision index.

A two-layer classification head, consisting of a linear layer, ReLU
activation, and a final linear layer, maps the state to class logits.
The class distribution is
\(\mathbf{p}_{i,k}^{(b)} = {softmax}\left( \mathcal{D}_{b}\left( \mathbf{h}_{i,k}^{(b)} \right) \right)\).
Stage 1 supervises every branch at every prefix using focal loss
\cite{lin2017focal}. With \(\Theta\) denoting all Stage-1
parameters, the objective is
\(\mathcal{L}_{Stage1}(\Theta) =\allowbreak \frac{1}{N}\sum_{i = 1}^{N}{\sum_{k = 1}^{K_{i}}{\sum_{b \in \mathcal{B}}^{}\mathcal{l}_{FL}}}\allowbreak\left( \mathbf{p}_{i,k}^{(b)},y_{i} \right)\).
The three branch losses have equal weight. Prefix supervision trains
each classifier to make predictions from partial observations, providing
the controller with prediction pathways usable before the complete
window is available.

\subsection{Joint Stopping and Branch
Selection}\label{joint-stopping-and-branch-selection}

After Stage 1, all feature-extraction, fusion, temporal-encoding, and
classification parameters are frozen. At each decision boundary, a
single controller receives the concatenated branch states in a fixed
order:

\begin{equation}
\mathbf{s}_{i,k} = \left\lbrack \mathbf{h}_{i,k}^{(F)}\|\mathbf{h}_{i,k}^{(1)}\|\mathbf{h}_{i,k}^{(2)} \right\rbrack \in \mathbb{R}^{3d}
\end{equation}

Class probabilities, entropy, observation fraction, and previous actions
are not explicitly appended to this state. Observation progress can
nevertheless influence the supplied states through the progress gate and
the prefix-specific projections \(\mathbf{W}_{b,k}^{P}\).

The controller is a linear mapping followed by deterministic action
selection:

\begin{equation}
\mathbf{u}_{i,k} = \mathbf{W}_{c}\mathbf{s}_{i,k} + \mathbf{b}_{c} \in \mathbb{R}^{4},a_{i,k} = \underset{a \in \left\{ 0,1,2,3 \right\}}{\arg\max}\mspace{6mu} u_{i,k}^{(a)}
\end{equation}

Actions 0, 1, 2, and 3 correspond to Continue, Stop-Fusion, Stop-MD1,
and Stop-MD2, respectively. The action scores are used directly, without
a softmax or action sampling. Continue advances to the next synchronized
snippet; each stopping action identifies both the stopping boundary and
the stopping boundary and the branch supplying the final prediction.

Let \(k_{i}^{*}\) be the first decision index with a non-Continue
action, and let \({\widehat{b}}_{i}\) be its associated branch. If the
controller selects Continue at every boundary, COMPASS terminates
at \(k_{i}^{*} = K_{i}\) and uses Fusion. The positive-class score is
read from the selected branch at that boundary:

\begin{equation}
q_{i} = p_{i,k_{i}^{*}}^{\left( {\widehat{b}}_{i} \right)}(y = 1),{\widehat{y}}_{i} = \mathbb{1}\left\lbrack q_{i} \geq \gamma^{*} \right\rbrack
\end{equation}

The threshold \(\gamma^{*}\) is determined during policy optimization
and retained for test inference. Both modalities are observed up to the
selected boundary; the terminal action selects the prediction source
rather than an independent sensor acquisition schedule.

\subsection{Multi-Objective Controller
Optimization}\label{multi-objective-controller-optimization}

We adopt the knee-guided neuroevolution algorithm introduced in SPN-V2
\cite{huang2024spnv2} to optimize the controller parameters
\(\theta = \left\{ \mathbf{W}_{c},\mathbf{b}_{c} \right\}\). Candidate
policies are evaluated using complete decision trajectories on the
training samples while Stage 1 remains fixed. The three minimized
objectives are
\(\mathbf{J}(\theta) = \left\lbrack 1 - {ACC}(\theta),\mspace{6mu}{EAR}_{train}(\theta),\mspace{6mu} 1 - {AUROC}(\theta) \right\rbrack\).
For each candidate, the selected branches produce continuous
positive-class scores. Its classification threshold is fitted by
maximizing Youden's \(J\) on the training ROC curve:

\begin{equation}
\gamma_{\theta}^{*} = \underset{\gamma}{\arg\max}\mspace{6mu}\left\lbrack {TPR}_{\theta}(\gamma) - {FPR}_{\theta}(\gamma) \right\rbrack
\end{equation}

Training accuracy uses this candidate-specific threshold, whereas AUROC
uses the continuous scores. The selected policy retains its associated
testing threshold.

The earliness objective follows the implementation's zero-based stopping
index. Expressed using our one-based decision notation, it is
\({EAR}_{train}(\theta) = \frac{1}{N}\sum_{i = 1}^{N}\frac{k_{i}^{*}(\theta) - 1}{K_{i}}\).
It assigns zero to stopping at the first snippet and remains below one
at the last snippet. For evaluation, earliness instead measures the
fraction of data actually observed. Using \(t_{i,k}\) from Section~\ref{problem-formulation}, the reported quantity is
\(\frac{1}{N}\sum_{i}^{}t_{i,k_{i}^{*}}/T_{i}\). For equal-duration
snippets covering the evaluated window, this becomes
\(\frac{1}{N}\sum_{i}^{}k_{i}^{*}/K_{i}\). The evolutionary search
generates mirrored perturbations of controller parameters:
\(\theta_{j}^{\pm} = \theta \pm \eta\mathbf{\epsilon}_{j},\mathbf{\epsilon}_{j} \sim \mathcal{N}\left( \mathbf{0},\mathbf{I} \right),j = 1,\ldots,50,\)
with perturbation scale \(\eta = 0.03\). Each iteration therefore
generates 50 mirrored pairs, or 100 perturbed candidates. Non-dominated
sorting compares their objective vectors. Objective-wise min--max
normalization produces costs \({\widetilde{J}}_{m}\), which are
evaluated using the achievement-scalarizing function
\(\Phi(\theta) = \max_{m \in \left\{ 1,2,3 \right\}}\left( w_{m}{\widetilde{J}}_{m}(\theta) \right) + \beta\sum_{m = 1}^{3}w_{m}{\widetilde{J}}_{m}(\theta),\)
where the objective weights are equal and \(\beta = 0.05\). The
lowest-scoring candidate provides a balanced compromise among the
normalized objectives. This candidate and the objective-wise best
candidates are retained to guide subsequent population generation and
updating. The search runs for 100 iterations, and the selected
compromise policy and its threshold form the inference model. Controller
learning uses these trajectory-level evolutionary evaluations rather
than step-wise reward-gradient updates.

\section{Experiments}\label{experiments}

We compare COMPASS with ETSC baselines, examine stopping
configurations and model components, and use a matched-stop analysis to
separate the effect of terminal branch selection from that of
observation time.

\subsection{Experimental Setup}\label{experimental-setup}

\textbf{Datasets and protocol.} We use six physiological and wearable
datasets \cite{penzel2000apnea,ichimaru1999psg,khalighi2016isruc,lehman2023vtac,schmidt2018wesad,physionet2007ucddb} (Table~\ref{tab:datasets}): four apnea tasks,
stress detection (WESAD), and true-versus-false ventricular-tachycardia
alarms (VTaC). All results use five-fold, segment-level stratified
cross-validation, reported as means and standard deviations across fold.

\begin{table}[!htbp]
\centering
\caption{Dataset configurations and binary classification tasks.}\label{tab:datasets}
\setlength{\tabcolsep}{1pt}
\begin{tabular*}{\linewidth}{@{\extracolsep{\fill}}lllrrr@{}}
\toprule
\textbf{Dataset} & \textbf{Target} & \textbf{Selected modalities} & \textbf{Window (s)} & \emph{\textbf{N\textsubscript{eval}}} & \textbf{Positive (\%)} \\
\midrule
MIT-BIH PSG & Apnea & ECG + Resp N& 30 & 8,867 & 34.4 (+) \\
Apnea-ECG & Apnea & ECG + Resp N& 60 & 3,409 & 29.4 (+) \\
UCDDB & Apnea & ECG + Resp N& 60 & 8,744 & 11.5 (+) \\
ISRUC-Sleep & Apnea & ECG + Resp N& 30 & 35,267 & 4.9 (+) \\
WESAD & Stress & ACC + BVP & 60 & 513 & 30.0 (+) \\
VTaC & VT alarm & ECG + ABP & 10 & 1,567 & 31.5 (+) \\
\bottomrule
\end{tabular*}
\par\vspace{2pt}
\begin{minipage}{\linewidth}
\raggedright
\fontsize{8}{9.5}\selectfont
\emph{N\textsubscript{eval}}: windows evaluated across the five test
folds; Positive: proportion of positive windows after preprocessing.
Window lengths refer to the full observation horizon. ECG: electrocardiogram;
Resp N: Nasal Respiration signal; ACC: acceleration; BVP: blood volume pulse;
ABP: arterial blood pressure. For VTaC, the observation window covers
the 10 s preceding alarm onset.
\end{minipage}
\end{table}

\textbf{Compared methods.} We compare COMPASS with multimodal
adaptations (MM) of five ETSC methods: TEASER \cite{schafer2020teaser}, ELECTS
\cite{russwurm2023elects}, SPN-V2 \cite{huang2024spnv2}, CALIMERA \cite{bilski2023calimera}, and UTSC
\cite{yan2025utsc}. Each baseline uses a single input configuration across all
metrics. We do not include Cao et al. \cite{cao2023early}, whose model is
designed for sequentially arriving, modality-tagged elements rather than
synchronized waveforms; adapting it would require redefining its
encoders, decision points, and waiting cost, which we have not done.
Internal comparisons reuse the frozen Stage 1 model. Fixed-fraction
stopping selects, on each training fold, the branch and snippet index
with the highest HM-ACC and applies them to every test sample.
Confidence-threshold stopping instead terminates when the selected
branch's maximum class probability first exceeds a threshold; the branch
and the threshold, searched over {[}0.5, 1{]}, are selected on training
data by HM-ACC in the same way. The multi-agent variant replaces
the single controller with three branch-specific agents, each mapping
its own branch state to Continue and Stop probabilities. A sample stops
when at least one agent selects Stop, and the prediction comes from the
branch whose agent assigns the highest Stop probability.

\textbf{Metrics.} We report HM-ACC, AUROC, sensitivity (SE), and
earliness, \(EAR = \frac{1}{N_{f}}\sum_{i}^{}\tau_{i}/T_{i}\), the
observed fraction of each window.
\(HM\text{-}ACC = 2\, ACC(1 - EAR)/\allowbreak(ACC + 1 - EAR)\) is computed within
each fold before averaging. COMPASS uses the threshold
\(\gamma^{*}\) selected on training data; All methods, including the
baselines, are classified with a threshold selected by
\emph{Youden's J} on training data.

\begin{table}[!htbp]
\centering
\caption{Comparison with multimodal (MM) adaptations of five ETSC
baselines (mean \textpm{} SD over five folds). \ensuremath{\uparrow}/\ensuremath{\downarrow}: higher/lower is better.
Bold/underline: best/second-best mean per dataset.}\label{tab:comparison}
\fontsize{9}{11}\selectfont
\setlength{\tabcolsep}{1pt}
\begin{tabular*}{\linewidth}{@{\extracolsep{\fill}}llrrrr@{}}
\toprule
\textbf{Dataset} & \textbf{Method} & \textbf{HM-ACC \ensuremath{\uparrow}} & \textbf{AUROC \ensuremath{\uparrow}} & \textbf{EAR \ensuremath{\downarrow}} & \textbf{SE \ensuremath{\uparrow}} \\
\midrule
\addlinespace[2pt]
\multirow{6}{*}{\textbf{MIT-BIH PSG}} & TEASER-MM & 0.8108\,\textpm\,0.0090 &
0.7381\,\textpm\,0.0126 & 0.0571\,\textpm\,0.0013 & 0.5839\,\textpm\,0.0453 \\
& ELECTS-MM & \ul{0.8726\,\textpm\,0.0039} & 0.8003\,\textpm\,0.0106 & \ul{0.0528\,\textpm\,0.0049} & 0.4329\,\textpm\,0.0351 \\
& SPN-V2-MM & 0.8086\,\textpm\,0.0205 & 0.8060\,\textpm\,0.0094 & 0.0779\,\textpm\,0.0071 &
0.7265\,\textpm\,0.0484 \\
& CALIMERA-MM & 0.8695\,\textpm\,0.0028 & \ul{0.8561\,\textpm\,0.0088} & 0.1008\,\textpm\,0.0018
& 0.7386\,\textpm\,0.0315 \\
& UTSC-MM & 0.8600\,\textpm\,0.0053 & 0.8276\,\textpm\,0.0046 & 0.0806\,\textpm\,0.0011 &
\textbf{0.7912\,\textpm\,0.0075} \\
& \textbf{COMPASS (Ours)} & \textbf{0.8739\,\textpm\,0.0017} & \textbf{0.8670\,\textpm\,0.0089} & \textbf{0.0354\,\textpm\,0.0009} & \ul{0.7428\,\textpm\,0.0176} \\
\addlinespace[2pt]
\multirow{6}{*}{\textbf{Apnea-ECG}} & TEASER-MM & 0.9152\,\textpm\,0.0073 &
0.9523\,\textpm\,0.0057 & 0.0607\,\textpm\,0.0058 & 0.8576\,\textpm\,0.0307 \\
& ELECTS-MM & 0.7658\,\textpm\,0.0079 & 0.9338\,\textpm\,0.0162 & 0.3468\,\textpm\,0.0138 &
0.8720\,\textpm\,0.0333 \\
& SPN-V2-MM & \ul{0.9175\,\textpm\,0.0048} & 0.9664\,\textpm\,0.0114 & 0.0933\,\textpm\,0.0060 &
\ul{0.8960\,\textpm\,0.0249} \\
& CALIMERA-MM & 0.8840\,\textpm\,0.0051 & \ul{0.9861\,\textpm\,0.0040} & 0.0687\,\textpm\,0.0021
& 0.5499\,\textpm\,0.0332 \\
& UTSC-MM & 0.8848\,\textpm\,0.0071 & 0.9796\,\textpm\,0.0067 & \ul{0.0457\,\textpm\,0.0001} &
0.5030\,\textpm\,0.0327 \\
& \textbf{COMPASS (Ours)} & \textbf{0.9543\,\textpm\,0.0042} & \textbf{0.9863\,\textpm\,0.0030} & \textbf{0.0446\,\textpm\,0.0014} & \textbf{0.9326\,\textpm\,0.0200} \\
\addlinespace[2pt]
\multirow{6}{*}{\textbf{UCDDB}} & TEASER-MM & 0.5792\,\textpm\,0.0047 &
\textbf{0.7929\,\textpm\,0.0308} & 0.5449\,\textpm\,0.0014 & \textbf{0.5851\,\textpm\,0.0591} \\
& ELECTS-MM & \textbf{0.8982\,\textpm\,0.0034} & 0.6633\,\textpm\,0.0175 & 0.0790\,\textpm\,0.0033 & 0.0752\,\textpm\,0.0151 \\
& SPN-V2-MM & 0.7667\,\textpm\,0.0541 & 0.6560\,\textpm\,0.0499 & 0.1692\,\textpm\,0.0168 &
\ul{0.4990\,\textpm\,0.1235} \\
& CALIMERA-MM & \ul{0.8964\,\textpm\,0.0029} & 0.6922\,\textpm\,0.0241 & 0.0978\,\textpm\,0.0019
& 0.3366\,\textpm\,0.0221 \\
& UTSC-MM & 0.7710\,\textpm\,0.1738 & 0.5107\,\textpm\,0.0132 & \textbf{0.0229\,\textpm\,0.0000}
& 0.3475\,\textpm\,0.3780 \\
& \textbf{COMPASS (Ours)} & 0.8772\,\textpm\,0.0085 & \ul{0.7186\,\textpm\,0.0302} &
\ul{0.0667\,\textpm\,0.0001} & 0.4673\,\textpm\,0.0645 \\
\addlinespace[2pt]
\multirow{6}{*}{\textbf{ISRUC-Sleep}} & TEASER-MM & 0.8509\,\textpm\,0.0094 &
0.6561\,\textpm\,0.0051 & \textbf{0.0601\,\textpm\,0.0003} & 0.4176\,\textpm\,0.0331 \\
& ELECTS-MM & \textbf{0.9301\,\textpm\,0.0019} & \textbf{0.8108\,\textpm\,0.0183} &
\ul{0.0948\,\textpm\,0.0018} & 0.2609\,\textpm\,0.0191 \\
& SPN-V2-MM & 0.3072\,\textpm\,0.1216 & 0.4612\,\textpm\,0.1483 & 0.6872\,\textpm\,0.0931 &
\textbf{0.6653\,\textpm\,0.3433} \\
& CALIMERA-MM & 0.7466\,\textpm\,0.0149 & \ul{0.8046\,\textpm\,0.0060} & 0.3291\,\textpm\,0.0130
& \ul{0.5443\,\textpm\,0.0373} \\
& UTSC-MM & 0.5307\,\textpm\,0.3742 & 0.5869\,\textpm\,0.1886 & 0.2133\,\textpm\,0.3812 & 0.5141\,\textpm\,0.4206 \\
& \textbf{COMPASS (Ours)} & \ul{0.9155\,\textpm\,0.0059} & 0.7438\,\textpm\,0.0202 &
0.0958\,\textpm\,0.0055 & 0.3554\,\textpm\,0.0377 \\
\addlinespace[2pt]
\multirow{6}{*}{\textbf{WESAD}} & TEASER-MM & 0.8319\,\textpm\,0.0188 & 0.7743\,\textpm\,0.0491 & \ul{0.0775\,\textpm\,0.0071} & 0.4022\,\textpm\,0.0542 \\
& ELECTS-MM & 0.7187\,\textpm\,0.1122 & 0.8839\,\textpm\,0.0798 & 0.3822\,\textpm\,0.1242 &
0.8095\,\textpm\,0.0903 \\
& SPN-V2-MM & \ul{0.8860\,\textpm\,0.0554} & \ul{0.9587\,\textpm\,0.0066} & 0.1251\,\textpm\,0.1007 & \ul{0.8284\,\textpm\,0.0750} \\
& CALIMERA-MM & 0.8174\,\textpm\,0.0142 & 0.9485\,\textpm\,0.0149 & 0.1262\,\textpm\,0.0090 &
0.2237\,\textpm\,0.0670 \\
& UTSC-MM & 0.8122\,\textpm\,0.0217 & 0.9443\,\textpm\,0.0315 & 0.1116\,\textpm\,0.0072 & 0.1581\,\textpm\,0.0960 \\
& \textbf{COMPASS (Ours)} & \textbf{0.9435\,\textpm\,0.0099} &
\textbf{0.9630\,\textpm\,0.0138} & \textbf{0.0360\,\textpm\,0.0033} & \textbf{0.8824\,\textpm\,0.0660} \\
\addlinespace[2pt]
\multirow{6}{*}{\textbf{VTaC}} & TEASER-MM & 0.7474\,\textpm\,0.0115 & 0.5798\,\textpm\,0.0251 & 0.0624\,\textpm\,0.0019 & 0.3255\,\textpm\,0.0335 \\
& ELECTS-MM & 0.7100\,\textpm\,0.0457 & \textbf{0.7401\,\textpm\,0.0322} & 0.3082\,\textpm\,0.0845 & 0.4844\,\textpm\,0.0577 \\
& SPN-V2-MM & 0.5478\,\textpm\,0.0920 & 0.6769\,\textpm\,0.1167 & 0.4398\,\textpm\,0.2534 &
\ul{0.5389\,\textpm\,0.0962} \\
& CALIMERA-MM & 0.7835\,\textpm\,0.0074 & 0.7214\,\textpm\,0.0115 & 0.1028\,\textpm\,0.0049 &
0.1958\,\textpm\,0.0494 \\
& UTSC-MM & \ul{0.7873\,\textpm\,0.0061} & 0.6396\,\textpm\,0.0295 & \ul{0.0557\,\textpm\,0.0134} & 0.1280\,\textpm\,0.0739 \\
& \textbf{COMPASS (Ours)} & \textbf{0.8108\,\textpm\,0.0079} & \ul{0.7226\,\textpm\,0.0242} & \textbf{0.0532\,\textpm\,0.0011} & \textbf{0.5853\,\textpm\,0.0272} \\
\bottomrule
\end{tabular*}
\end{table}

\subsection{Overall Comparison}\label{overall-comparison}

COMPASS with the best average ranks across the six datasets in
HM-ACC (1.50), AUROC (1.67), and EAR (1.50) (Table~\ref{tab:comparison}). It has the
highest HM-ACC and the lowest EAR on MIT-BIH PSG, Apnea-ECG, WESAD, and
VTaC, with the largest margins on Apnea-ECG (0.9543 vs. 0.9175 for
SPN-V2-MM) and WESAD (0.9435 vs. 0.8860).

COMPASS does not lead on every dataset. ELECTS-MM has higher HM-ACC
and AUROC on ISRUC-Sleep. On UCDDB, ELECTS-MM and CALIMERA-MM have
higher HM-ACC but lower AUROC and SE, whereas TEASER-MM attains the
highest AUROC and SE while observing 0.5449 of the window, compared with
0.0667 for COMPASS. Because accuracy-based scores can mask weak
positive-class detection, we also examine sensitivity on UCDDB. For example,
an ECG-only UTSC configuration reaches an HM-ACC of 0.9185 with an SE
of only 0.0277.

\subsection{Stopping Configurations and
Components}\label{stopping-configurations-and-components}

Table~\ref{tab:stopping} compares complete stopping configurations. The learned
policy yields the highest HM-ACC on four datasets, most clearly on
MIT-BIH PSG (0.8739 vs. 0.8223 and 0.8287), where it also stops earlier.
Fixed-fraction stopping is marginally better on Apnea-ECG and WESAD
and often gives higher AUROC and SE. Learning the decision configuration
thus improves the accuracy--earliness balance without improving every
criterion.

\begin{table}[!htbp]
\centering
\caption{Comparison of stopping configurations (mean \textpm{} SD over
five folds). Bold: best mean per dataset.}\label{tab:stopping}
\fontsize{9}{11}\selectfont
\setlength{\tabcolsep}{1pt}
\begin{tabular*}{\linewidth}{@{\extracolsep{\fill}}llrrrr@{}}
\toprule
\textbf{Dataset} & \textbf{Stopping configuration} & \textbf{HM-ACC \ensuremath{\uparrow}} & \textbf{AUROC \ensuremath{\uparrow}} & \textbf{EAR \ensuremath{\downarrow}} & \textbf{SE \ensuremath{\uparrow}} \\
\midrule
\addlinespace[2pt]
\multirow{3}{*}{\textbf{MIT-BIH PSG}} & Fixed fraction & 0.8223\,\textpm\,0.0433
& 0.7737\,\textpm\,0.0742 & 0.0965\,\textpm\,0.0551 & 0.6169\,\textpm\,0.1057 \\
& Confidence threshold & 0.8287\,\textpm\,0.0124 & 0.6454\,\textpm\,0.0195 & 0.1220\,\textpm\,0.0144 & 0.3411\,\textpm\,0.0330 \\
& \textbf{COMPASS (Ours)} & \textbf{0.8739\,\textpm\,0.0017} & \textbf{0.8670\,\textpm\,0.0089} & \textbf{0.0354\,\textpm\,0.0009} & \textbf{0.7428\,\textpm\,0.0176} \\
\addlinespace[2pt]
\multirow{3}{*}{\textbf{Apnea-ECG}} & Fixed fraction & \textbf{0.9546\,\textpm\,0.0029} & 0.9854\,\textpm\,0.0030 & \textbf{0.0434\,\textpm\,0.0000} & \textbf{0.9420\,\textpm\,0.0133} \\
& Confidence threshold & 0.9508\,\textpm\,0.0025 & 0.9833\,\textpm\,0.0038 & 0.0524\,\textpm\,0.0059 & 0.9232\,\textpm\,0.0217 \\
& \textbf{COMPASS (Ours)} & 0.9543\,\textpm\,0.0042 & \textbf{0.9863\,\textpm\,0.0030} &
0.0446\,\textpm\,0.0014 & 0.9326\,\textpm\,0.0200 \\
\addlinespace[2pt]
\multirow{3}{*}{\textbf{UCDDB}} & Fixed fraction & 0.8613\,\textpm\,0.0138 &
\textbf{0.7556\,\textpm\,0.0157} & 0.0711\,\textpm\,0.0000 & \textbf{0.5218\,\textpm\,0.0627} \\
& Confidence threshold & 0.8635\,\textpm\,0.0166 & 0.7523\,\textpm\,0.0141 & 0.0712\,\textpm\,0.0001 & 0.4980\,\textpm\,0.0988 \\
& \textbf{COMPASS (Ours)} & \textbf{0.8772\,\textpm\,0.0085} & 0.7186\,\textpm\,0.0302 &
\textbf{0.0667\,\textpm\,0.0001} & 0.4673\,\textpm\,0.0645 \\
\addlinespace[2pt]
\multirow{3}{*}{\textbf{ISRUC-Sleep}} & Fixed fraction & 0.8920\,\textpm\,0.0047
& \textbf{0.7560\,\textpm\,0.0089} & \textbf{0.0909\,\textpm\,0.0000} & \textbf{0.4631\,\textpm\,0.0127} \\
& Confidence threshold & 0.8977\,\textpm\,0.0107 & 0.7381\,\textpm\,0.0113 & 0.1354\,\textpm\,0.0151 & 0.3703\,\textpm\,0.0215 \\
& \textbf{COMPASS (Ours)} & \textbf{0.9155\,\textpm\,0.0059} & 0.7438\,\textpm\,0.0202 &
0.0958\,\textpm\,0.0055 & 0.3554\,\textpm\,0.0377 \\
\addlinespace[2pt]
\multirow{3}{*}{\textbf{WESAD}} & Fixed fraction & \textbf{0.9447\,\textpm\,0.0089} & \textbf{0.9667\,\textpm\,0.0154} & \textbf{0.0333\,\textpm\,0.0000} &
\textbf{0.8883\,\textpm\,0.0326} \\
& Confidence threshold & 0.9413\,\textpm\,0.0100 & 0.9590\,\textpm\,0.0156 & 0.0402\,\textpm\,0.0090 & 0.8621\,\textpm\,0.0226 \\
& \textbf{COMPASS (Ours)} & 0.9364\,\textpm\,0.0071 & 0.9630\,\textpm\,0.0138 &
0.0360\,\textpm\,0.0033 & 0.8824\,\textpm\,0.0660 \\
\addlinespace[2pt]
\multirow{3}{*}{\textbf{VTaC}} & Fixed fraction & 0.8017\,\textpm\,0.0114 &
\textbf{0.7368\,\textpm\,0.0263} & 0.0736\,\textpm\,0.0257 & \textbf{0.6182\,\textpm\,0.0571} \\
& Confidence threshold & 0.8094\,\textpm\,0.0086 & 0.7136\,\textpm\,0.0442 & 0.0849\,\textpm\,0.0250 & 0.5178\,\textpm\,0.1072 \\
& \textbf{COMPASS (Ours)} & \textbf{0.8108\,\textpm\,0.0079} & 0.7226\,\textpm\,0.0242 &
\textbf{0.0532\,\textpm\,0.0011} & 0.5853\,\textpm\,0.0272 \\
\bottomrule
\end{tabular*}
\end{table}

In the ablations (Table~\ref{tab:ablation}), the single-controller approach
outperforms the multi-agent controller across all six datasets by
0.08--4.50 percentage points. Removing gated attention or the feature
adapters lowers HM-ACC by up to 1.49 and 3.71 points, respectively, and
replacing the Transformer with an LSTM causes the largest drops, up to
31.69 points on VTaC. Representation learning therefore contributes
substantially, and the overall gains should not be attributed solely to
branch selection.

\begin{table}[!htbp]
\centering
\caption{Component ablations. Full: HM-ACC of the full model (mean
\textpm{} SD). Other columns: full model minus variant, in percentage points.
Positive values favor the full model.}\label{tab:ablation}
\fontsize{9}{11}\selectfont
\setlength{\tabcolsep}{1pt}
\begin{tabular*}{\linewidth}{@{\extracolsep{\fill}}lrrrrr@{}}
\toprule
\textbf{Dataset} & \textbf{Full HM-ACC \ensuremath{\uparrow}} & \textbf{vs. multi-agent (pp)} & \textbf{vs. no gate (pp)} & \textbf{vs. no adapter (pp)} & \textbf{vs. LSTM (pp)} \\
\midrule
MIT-BIH PSG & 0.8739 \textpm{} 0.0017 & +1.95 & +0.22 & +0.29 & +7.89 \\
Apnea-ECG & 0.9543 \textpm{} 0.0042 & +0.08 & +0.35 & +0.10 & +3.33 \\
UCDDB & 0.8772 \textpm{} 0.0085 & +4.50 & +0.51 & +0.65 & +20.45 \\
ISRUC-Sleep & 0.9155 \textpm{} 0.0059 & +1.98 & +0.86 & +0.65 & +3.60 \\
WESAD & 0.9464 \textpm{} 0.0071 & +0.70 & +0.72 & +0.37 & +19.72 \\
VTaC & 0.8108 \textpm{} 0.0079 & +1.57 & +1.49 & +3.71 & +31.69 \\
\bottomrule
\end{tabular*}
\end{table}

\subsection{Matched-Stop Analysis of Branch
Selection}\label{matched-stop-analysis-of-branch-selection}

To isolate branch selection, we use the matched-stop fixed-fusion
readout control. For each test sample, we keep the stopping point
selected by COMPASS and compare its routed prediction, produced by
the selected branch, with the fusion branch's prediction at the same
point. A rescue occurs when only the routed prediction is correct, and a
harm when only the fusion prediction is correct. The net change for fold
\(f\) is
\(\Delta{ACC}_{f} = 100\,\left( n_{rescue,f} - n_{harm,f} \right)/N_{f}\)
percentage points, averaged equally over folds. This control does not
retrain a fusion-only controller with its own stopping times.

\begin{figure}[!htbp]
\centering
\includegraphics[width=\linewidth]{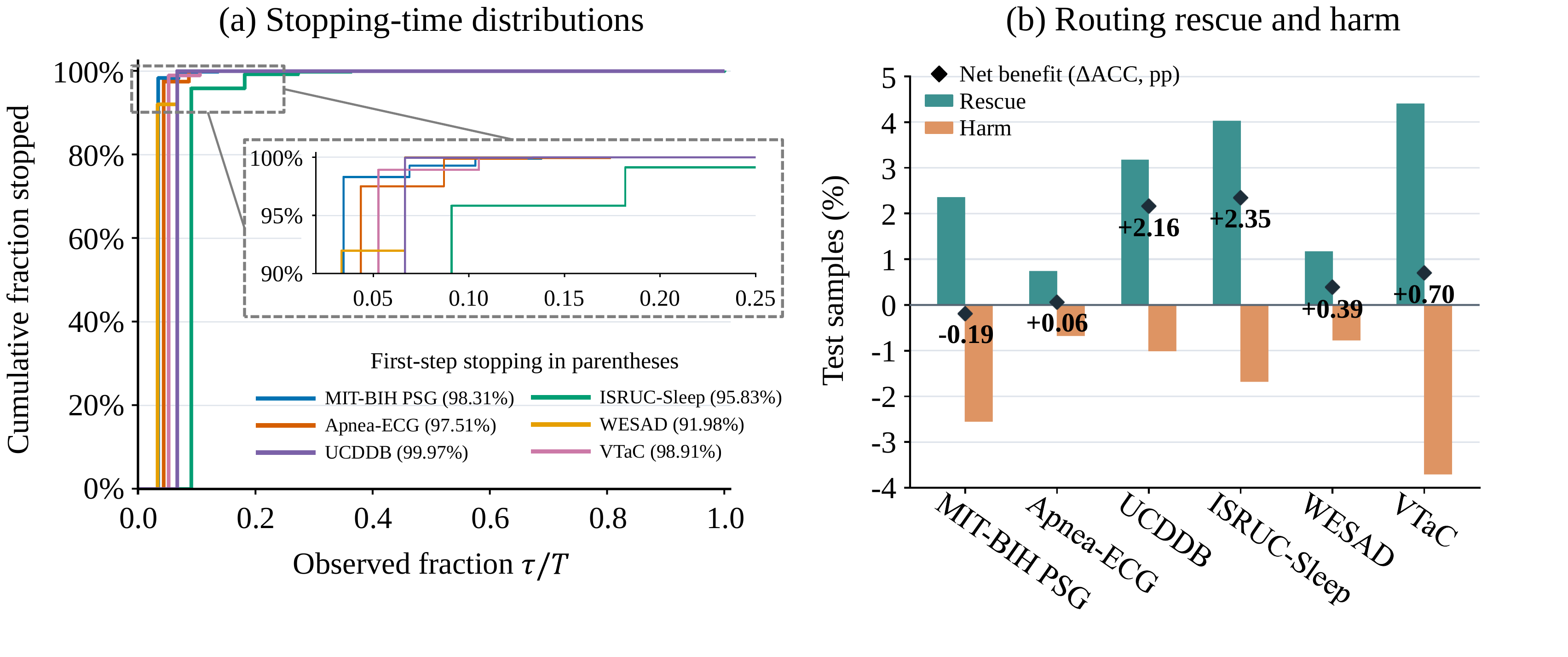}
\caption{(a) Cumulative stopping distributions averaged over five folds.
First-step stopping rates in parentheses. (b) Rescue and harm rates
under the matched-stop fixed-fusion readout control. Diamonds show the
net \ensuremath{\Delta}ACC in percentage points.}\label{fig:matched-stop}
\end{figure}

Most decisions occur at the first decision point, with first-step rates
ranging from 91.98\% to 99.97\% (Figure~\ref{fig:matched-stop}a). Low earliness
therefore reflects early commitment rather than extended observation,
and the branch chosen at that point determines most predictions.

\begin{figure}[!htbp]
\centering
\includegraphics[width=.78\linewidth]{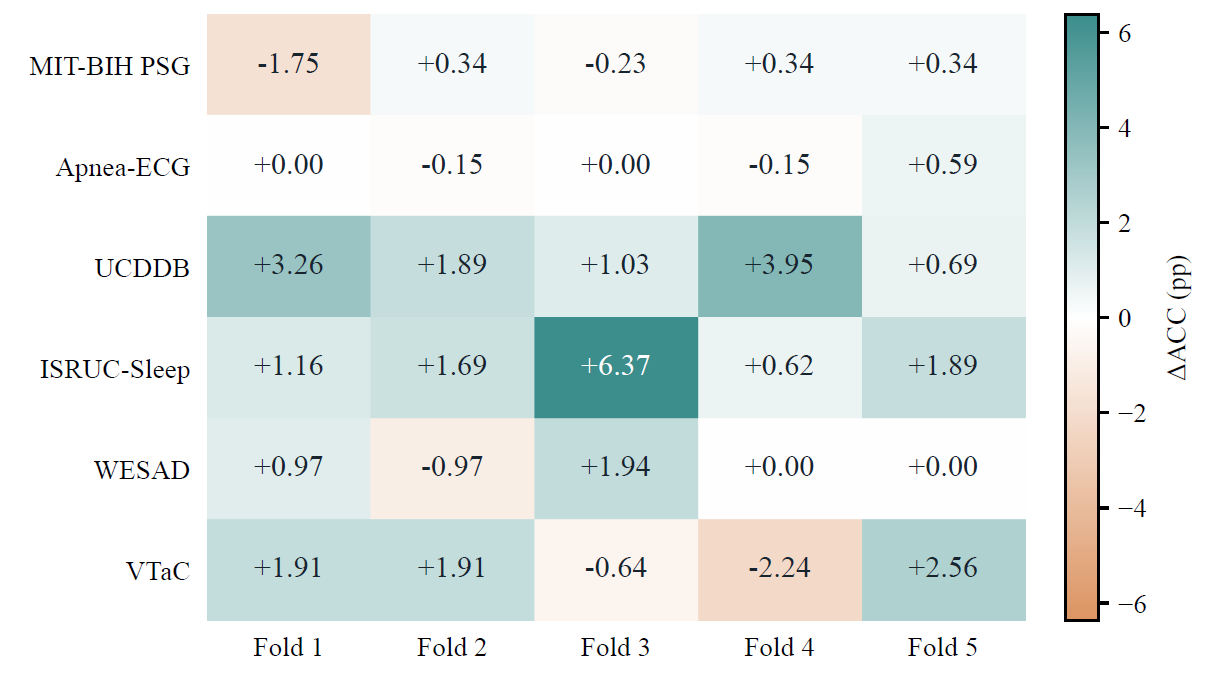}
\caption{Fold-level \ensuremath{\Delta}ACC (percentage points) under the
matched-stop fixed-fusion readout control. Positive values favor branch
selection.}\label{fig:fold-accuracy}
\end{figure}

Branch selection improves mean accuracy on five of the six datasets
(Figure~\ref{fig:matched-stop}b), most on ISRUC-Sleep (+2.35 points; rescue 4.02\%, harm
1.68\%) and UCDDB (+2.16 points), with smaller gains on VTaC, WESAD, and
Apnea-ECG. MIT-BIH PSG shows a small decrease (\textminus{}0.19 points; rescue:
2.36\%; harm: 2.55\%). Gains are positive in every fold of UCDDB and
ISRUC-Sleep but mixed elsewhere (Figure~\ref{fig:fold-accuracy}). These effects are
distinct from overall performance: COMPASS leads in HM-ACC on MIT-BIH
PSG despite a negative routing effect, whereas routing helps on UCDDB,
where COMPASS does not lead. Alternative branches can thus correct
fusion errors without additional observation, but the benefit is
dataset-dependent, and the controller does not always select the better
branch.

\FloatBarrier
\section{Conclusion}\label{conclusion}

We presented COMPASS, a multimodal early time-series classification
framework in which a single controller decides both when to stop and
which modality-associated or fused branch produces the prediction. On
six physiological and wearable datasets, it achieved the best average
accuracy-earliness ranks among the evaluated baselines. At matched
stopping times, branch selection improved accuracy over the fused
readout on five datasets, but with dataset-dependent harm. Because most
decisions occur at the first decision point, future work will examine
adaptive waiting, the interaction between the progress gate and branch
selection, and subject-independent generalization.

\bibliographystyle{splncs04-citation-order}
\bibliography{references}
\end{document}